\documentclass[journal]{IEEEtai}

\usepackage{enumerate}
\usepackage{cite}
\usepackage[pdftex]{graphicx}
\usepackage{amsmath}
\usepackage{array}
\usepackage{amssymb}
\usepackage{bm}
\usepackage{makecell}
\usepackage{multirow}
\usepackage{color}

\usepackage{algorithmic}
\usepackage{algorithm}

\newtheorem{definition}{Definition}
\newcommand{\eref}[1]{(\ref{#1})}
\newcommand{\fref}[1]{Fig.~\ref{#1}}
\newcommand{\sref}[1]{Sec.~\ref{#1}}

\newcommand{\Alref}[1]{Algorithm~\ref{#1}}

\makeatletter
\newcommand{\removelatexerror}{\let\@latex@error\@gobble}
\makeatother

\begin{document}

\title{Residual Tensor Train: A Quantum-inspired Approach for Learning Multiple Multilinear Correlations}
\author{Yiwei~Chen, 
        Yu~Pan,~\IEEEmembership{Senior~Member,~IEEE,}
        Daoyi~Dong,~\IEEEmembership{Senior~Member,~IEEE}
        
\thanks{Y. Chen is with the Institute of Cyber-Systems and Control, College of Control Science and Engineering, Zhejiang University, Hangzhou, 310027, China. (email: ewell@zju.edu.cn).}
\thanks{Y. Pan is with the State Key Laboratory of Industrial Control Technology, Institute of Cyber-Systems and Control, College of Control Science and Engineering, Zhejiang University, Hangzhou, 310027, China. (email: ypan@zju.edu.cn).}
\thanks{D. Dong is with the School of Engineering and Information
Technology, University of New South Wales, Canberra, ACT 2600,
Australia. (email: daoyidong@gmail.com).}

}

\maketitle

\begin{abstract}
States of quantum many-body systems are defined in a high-dimensional Hilbert space, where rich and complex interactions among subsystems can be modelled. In machine learning, complex multiple multilinear correlations may also exist within input features. In this paper, we present a quantum-inspired multilinear model, named Residual Tensor Train (ResTT), to capture the multiple multilinear correlations of features, from low to high orders, within a single model. ResTT is able to build a robust decision boundary in a high-dimensional space for solving fitting and classification tasks. In particular, we prove that the fully-connected layer and the Volterra series can be taken as special cases of ResTT. Furthermore, we derive the rule for weight initialization that stabilizes the training of ResTT based on a mean-field analysis. We prove that such a rule is much more relaxed than that of TT, which means ResTT can easily address the vanishing and exploding gradient problem that exists in the existing TT models. Numerical experiments demonstrate that ResTT outperforms the state-of-the-art tensor network and benchmark deep learning models on MNIST and Fashion-MNIST datasets.  Moreover, ResTT achieves better performance than other statistical methods on two practical examples with limited data which are known to have complex feature interactions.
\end{abstract}

\begin{IEEEImpStatement}
Optimal learning may require considering high-order multilinear feature interactions. Recently, quantum-inspired tensor networks have been proposed to exploit high-order correlations. However, the vanilla tensor networks are specialized in capturing only one type of multilinear feature interaction. This paper proposes a novel network design which is able to describe generic feature interactions of different orders, within one model. Due to the residual connections, the training stability has also been promoted to an unprecedented level when compared to previous tensor network methods. Our methodology offers an alternative way of building a robust and efficient model for data deficient tasks which may have complex feature interactions.
\end{IEEEImpStatement}


\section{Introduction}
\label{Introduction}

\IEEEPARstart{M}{ultilinear} feature correlations play an important role in machine learning \cite{lu2008mpca, lu2011survey, yang2017deep}. Linear methods, such as linear regression \cite{freedman2009statistical}, leverage linear correlation to solve simple machine learning problems. However, a single linear correlation may underfit the data that contain complex underlying patterns. In \cite{tenenbaum2000separating}, a bilinear model has been utilized to separate the style and content of handwritten digits. Since then, low-order feature interactions, including bilinear and trilinear correlations, have been widely adopted to improve the capability of modelling the multilinearity hidden in data \cite{lin2015bilinear,gao2016compact,zheng2019looking,do2019compact}. For high-order feature interactions, due to the dimensionality issue which leads to expensive computational and memory costs, the modelling of multilinear correlations is still a challenging problem. 

\begin{figure}
	\centering
	\includegraphics[width=0.38\textwidth]{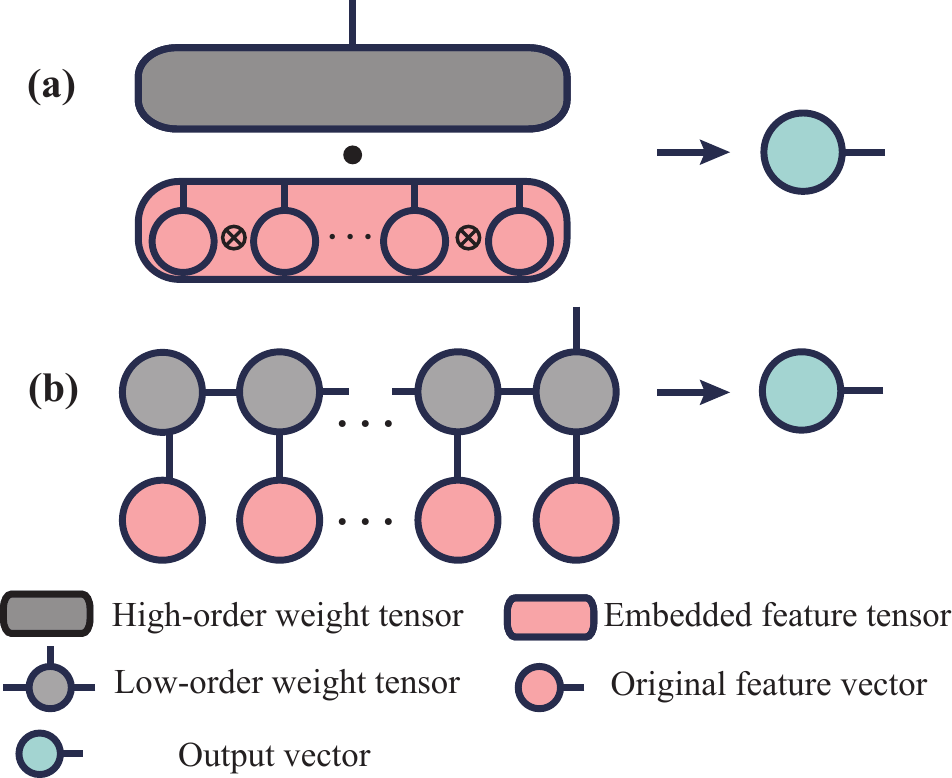}
	\caption{(a) The product of the embedded feature tensor and the high-order weight tensor. (b) The tensor network representation of the input features contracting with the high-order weight tensor. The high-order weight tensor is approximated by the one-dimensional TT. The contraction between the TT and input features generates an output vector.}
	\label{Fig:TT-learning}
\end{figure}

Recently, quantum-inspired machine learning has drawn much attention and demonstrated potential applicability in computer vision \cite{stoudenmire2016supervised, xu2020hyper, liu2020low-rank}, natural language processing \cite{sordoni2013modeling,kartsaklis2021lambeq, chen2021quantum}, recommendation system \cite{novikov2017exponential}. Increasing efforts \cite{han2018unsupervised,stoudenmire2016supervised,liu2018machine,efthymiou2019tensornetwork,huggins2018towards} have been made to build multilinear model based on tools that were originally developed for studying many-body quantum systems, such as tensor train (TT) \cite{bridgeman2017hand,evenbly2011tensor}. In the TT model, input features are embedded into a high-dimensional Hilbert space by tensor product, in which a simple decision boundary may correspond to a highly complex one in the original low-dimensional feature space. Thus, by defining a trainable high-order weight tensor, multilinear interactions can be modelled by simply multiplying the embedded input with this tensor (see \fref{Fig:TT-learning}(a)). In order to overcome the dimensionality issue, a one-dimensional chain of low-order tensors, namely TT, can be used to approximate the high-order weight tensor (see \fref{Fig:TT-learning}(b)) to significantly reduce the number of trainable parameters. Particularly, for $N$ input features, the vanilla TT model captures $N$-order multilinear correlations. Nevertheless, linear, bilinear and higher-order correlations may exist simultaneously among features, and thus a more generalized model needs to be built to cover these scenarios.

In this paper, we propose a Residual TT (ResTT) which is capable of modelling multiple multilinear feature interactions within a single model. The main innovation of ResTT is to incorporate skip connections \cite{he2016deep} into the present TT model. By designing the network topology of skip connections, the ResTT can flexibly model diverse types of feature interactions. The main advantages of ResTT and contributions of this paper are summarized as follows:

\begin{itemize}
	\item A novel multilinear model is proposed. ResTT is able to learn multiple multilinear feature correlations in a data-driven way. Besides, ResTT can flexibly model a diverse set of feature interactions by simply modifying the skip connections. In particular, by designing the network topology of skip connections, ResTT can be reduced to the fully-connected layer \cite{lecun2015deep} and Volterra series \cite{s1985fading}. In other words, the fully-connected layer and Volterra series can be looked as special cases of ResTT.
	
	\item The mean-field analysis \cite{saxe2013exact} is extended to TT-based models for deriving the rules of initialization that stabilize the training process. In particular, ResTT is demonstrated to be immune to the gradient vanishing and exploding problem with a much more relaxed training stability condition than that of the previous TT \cite{novikov2017exponential}. We find that the variance of input features will disturb the training of vanilla TT and a strict initial condition has to be satisfied to ensure its stability. In contrast, the stability condition for ResTT is significantly relaxed, and there is no need to impose strict restrictions on the statistics of the input features.
	
	\item Comprehensive experiments are conducted on the synthetic dataset, the image classification tasks, including MNIST \cite{lecun2010} and Fashion-MNIST \cite{xiao2017fashion}, and two practical tasks. The results on MNIST and Fashion-MNIST demonstrate that ResTT consistently outperforms the state-of-the-art tensor network models. In particular, the convergence of ResTT is faster and more stable than the plain TT. Inspired by the experimental setup in \cite{yang2017deep}, ResTT is also compared with the benchmark deep learning models on limited training data (1\%, 5\% and 10\%) and has demonstrated superior performance both in prediction accuracy and training robustness. We focus on this comparison since limited training data is reported as a major issue in many practical tasks, in which data acquisition is expensive and the lack of training samples may easily cause overfitting \cite{hu2017frankenstein}. Besides, ResTT achieves the best performance on Boston Housing \cite{harrison1978hedonic} and ALE \cite{singh2020machine} datasets as compared to other statistical algorithms, which further demonstrates its advantage in modelling complex feature interactions with limited training data.
\end{itemize}

The remainder of this paper is organized as follows. \sref{2} provides a brief introduction to the background and related works. In \sref{3}, we introduce ResTT in detail. \sref{4} presents the mean-field analysis for TT and ResTT. \sref{5} presents the experimental results on the synthetic and real datasets. Finally, the conclusion is drawn in \sref{6}.

\section{Preliminaries}
\label{2}
In this section, we first introduce the notation and definitions used in this paper and then we review the TT approach for supervised learning tasks \cite{stoudenmire2016supervised}. Lastly, the related works are discussed.
\subsection{Notation and definition}
\label{2-1}
\begin{definition}[Tensor]
	Tensor, also known as multidimensional or $N$-mode array, is a generalization of vector (one index) and matrix (two indices) to an arbitrary number of indices. In this paper, we denote tensors by calligraphic capitals. An $N$-order tensor is denoted by $\mathcal{A} \in \mathbb{R}^{I_{1} \times I_{2} \times \cdots \times I_{N}}$, in which $I_{n}$ ($1 \leq n\leq N, n \in \mathbb{N}^{+}$) is the dimension of the $n$-th index. The element of $\mathcal{A}$ is denoted by $a_{i_{1} i_{2} \cdots i_{N}}$ ($1\leq i_{n}\leq I_{n}, I_{n} \in \mathbb{N}^{+}$).
\end{definition}

\begin{figure}
 	\centering
 	\includegraphics[width=0.45\textwidth]{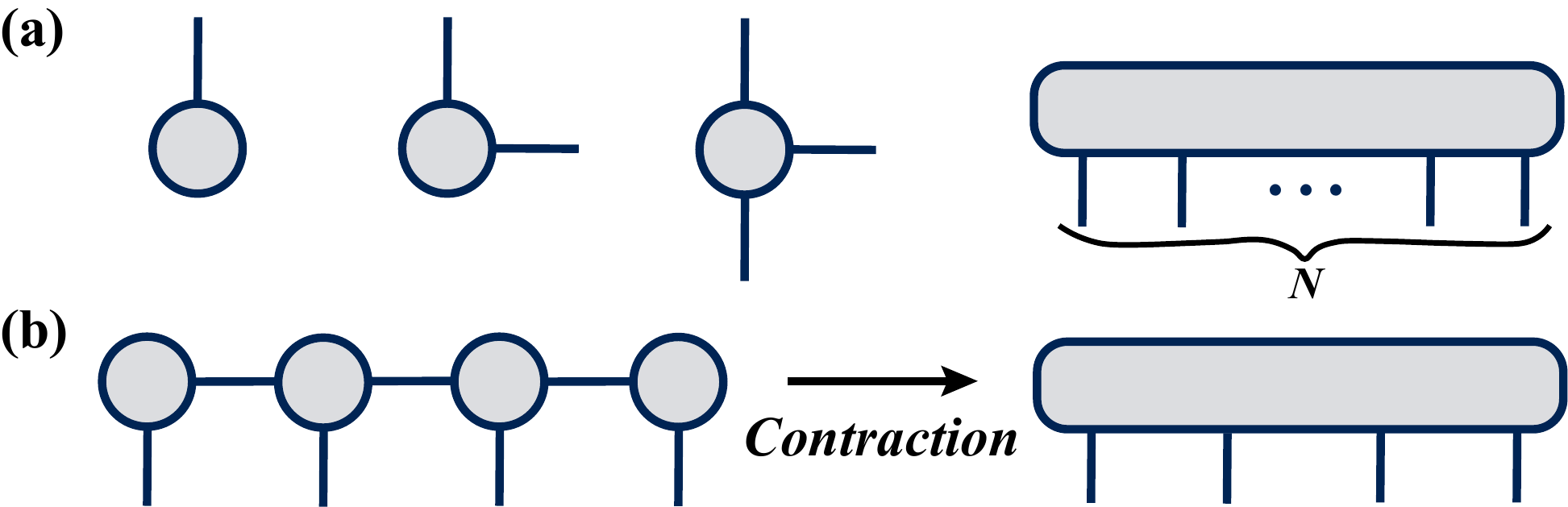}
 	\caption{Tensor network notation.
 		\textbf{(a)} Graphical notations of tensors with different orders.
 		\textbf{(b)} Contracting the four nodes in TT to obtain a 4-order tensor.}
 	\label{Fig:TN_notation}
 \end{figure}

The diagrammatic notation for a tensor is drawn as a circle with edges, in which the circle represents the elements of the tensor and the edges represent its each individual index. For example, the graphics shown in \fref{Fig:TN_notation} (a) represent vector ($1$-order tensor), matrix ($2$-order tensor), $3$-order tensor and $N$-order tensor, respectively.


\begin{definition}[Tensor Contraction]
	Tensor contraction is an operation that combines two or more tensors into a new one. In this paper, we denote tensor contraction by ${\rm C}_{I_{k}}[\cdot]$, where $I_{k}$ is the index to be contracted. For example, a matrix $\mathcal{M}=\{m_{i_{1}i_{2}}\} \in \mathbb{R}^{I_{1} \times I_{2}}$ and a 3-order tensor $\mathcal{T}\in \mathbb{R}^{J_{1} \times J_{2} \times I_{1}}$ can be contracted along the index $I_{1}$ as
	\begin{equation}
	\mathcal{C} = {\rm C}_{I_{1}}[\mathcal{M},\mathcal{T}]
	\label{Eq:tensor_contraction}
	\end{equation}
	with $\mathcal{C} \in \mathbb{R}^{I_{2} \times J_{1} \times J_{2}}$ being the new 3-order tensor. The elements of $\mathcal{C}$ are given by
	\begin{equation}
	c_{i_{2} j_{1}j_{2}}= \sum_{i_{1}}m_{i_{1}i_{2}}t_{j_{1}j_{2}i_{1}}. \label{Eq:tensor_contraction_element}
	\end{equation}	
Note that the tensors to be contracted must have one or more compatible indices of the same dimension, such as the index $I_{1}$ in $\mathcal{M}$ and $\mathcal{T}$.
\end{definition}

In the graphical notation, a contraction is represented by a sharing edge, which indicates that the two tensor nodes are contracted along this particular index.

\begin{definition}[Tensor Train]
	Tensor Train (TT) is a set of tensors to be contracted in a one-dimensional mode. Specifically, $N$ tensors denoted by $\{\mathcal{W}^{(1)} \in \mathbb{R}^{I_{1}\times V_{1}}, \mathcal{W}^{(2)} \in \mathbb{R}^{V_{1}\times I_{2} \times V_{2}}, \cdots, \mathcal{W}^{(N)} \in \mathbb{R}^{V_{N-1} \times I_{N}}\}$ constitute a TT if contracted as
	\begin{equation}
	\mathcal{W} = {\rm C}_{V_1,\cdots,V_{N-1}}[\mathcal{W}^{(1)},\mathcal{W}^{(2)},\cdots,\mathcal{W}^{(N)}],
	\label{Eq:tensor_train}
	\end{equation}
	where $\mathcal{W} \in \mathbb{R}^{I_{1} \times \cdots \times I_{N}}$ is the generated $N$-order tensor and $V_{1}, V_{2}, \cdots, V_{N-1}$ are called virtual bond dimensions.
\end{definition}

The graphical representation of the contraction process of a 4-node TT is drawn in \fref{Fig:TN_notation} (b). It is worth mentioning that the virtual bond dimensions could be used to control the total number of parameters.

\begin{definition}[Feature interaction]
	In statistics, feature interaction \cite{dodge2006oxford} describes a situation in which the effect of one feature on the outcome depends on the state of other features. It is common to use products of features, also known as multilinear feature correlations, to represent different types of feature interaction. The number of features in the product term is defined as the order of interaction. For example, a bilinear model with an output $y$ and two input features $x_1$ and $x_2$ can be formulated as
	\begin{equation}
	y = c \cdot x_{1}x_{2},
	\end{equation}
	where $c \cdot x_{1}x_{2}$ describes the 2-order interaction between $x_1$ and $x_2$.
\end{definition}

\subsection{Multilinear learning approaches}
\label{2-3}
Linear methods have a wide range of applications \cite{weisberg2005applied}. However, linear model may not be sufficient to describe the correlation between features for complicated tasks, and thus multilinear methods have attracted increasingly attention in recent years. Early studies of multilinear models mainly focused on bilinear \cite{tenenbaum2000separating} and trilinear correlations \cite{karihaloo2013determination}. Recently, low-order feature interactions have been adopted into deep learning models. The work in \cite{socher2013reasoning} utilized the bilinear correlation to construct a novel neural network layer, which provides a powerful way to describe relational information than the standard fully-connected layer. Combined with convolutional layers, such a layer has shown effectiveness in the question answering task \cite{qiu2015convolutional}. Based on this idea, novel neural networks modelling bi- or tri-linear correlations have been proposed and demonstrated good performance in image classification \cite{gao2016compact}, fine-grained image recognition \cite{lin2015bilinear, do2019compact} and visual question answering \cite{do2019compact}. Besides, the work in \cite{lin2016multi} proposed a bilinear model to exploit the task relatedness from feature interactions in multi-task learning.

Modelling high-order feature interactions is much more difficult than modelling low-order ones due to the dimensionality issue. For example, modelling the high-dimensional and noisy visual features in video semantic recognition usually suffers from the curse of dimensionality \cite{luo2017adaptive}. Some multilinear learning frameworks, such as multilinear principal component analysis \cite{lu2008mpca}, employ dimensionality reduction techniques to simplify the original high-order tensor and cannot be trained via an end-to-end way. Recently, the quantum-inspired TT approaches \cite{blondel2016polynomial,novikov2017exponential}, trained by gradient decent, were proposed to extract the $N$-order dependencies among $N$ input features on the recommendation system tasks. After that, more and more quantum-inspired approaches were proposed to use Matrix Product State (MPS) \cite{stoudenmire2016supervised,han2018unsupervised}, Tree Tensor Network (TTN) \cite{liu2018machine} and other types of low-dimensional tensor networks \cite{glasser2018supervised} on image classification and generation tasks. However, as indicated by \cite{novikov2017exponential} and \cite{liu2018machine}, the vanilla TT may fail to converge in training in case of a long chain, since hundreds of contractions will result in unbounded or vanishing outputs. Thus, specific normalization procedures must be embedded into the computation steps to keep the gradients in a reasonable range. To make things worse, the current TT lacks generalizability and flexibility since only the $N$-order feature interaction has been modelled, and thus the comprehensive multilinear correlations cannot be fully captured.

\subsection{Tensor Train Approach}
\label{2-2}
\begin{figure}
	\centering
	\includegraphics[width=0.32\textwidth]{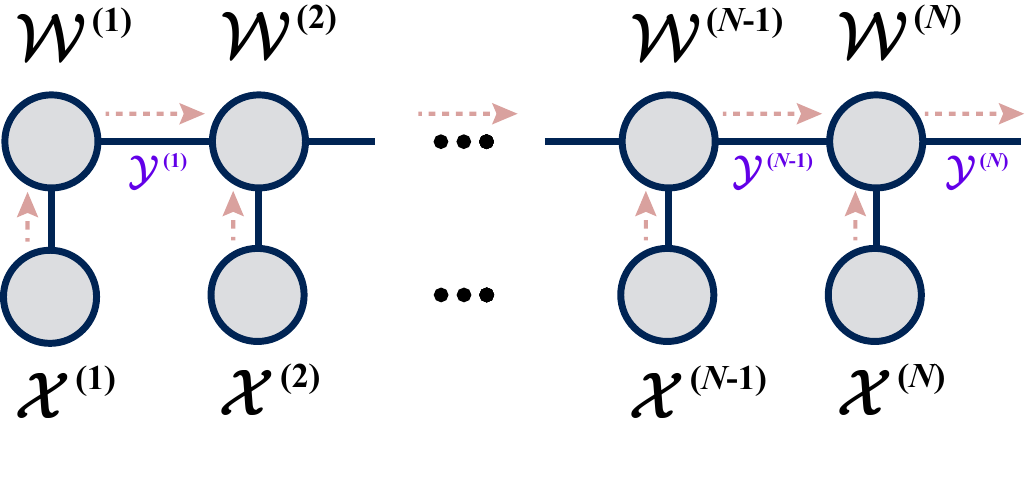}
	\caption{The TT model that is contracted from left to right. The outputs after the contractions are denoted in a blue font.}
	\label{Fig:TT-structure}
\end{figure}

TT approach \cite{stoudenmire2016supervised} provides an effective way to model high-order feature interactions. Given a set of feature vectors $ \{ \mathcal{X}^{(1)} \in \mathbb{R}^{I_{1}},  \mathcal{X}^{(2)} \in \mathbb{R}^{I_{2}},\cdots, \mathcal{X}^{(N)}\in \mathbb{R}^{I_{N}}\}$, the network model is defined as
\begin{equation}
\mathcal{Y} = {\rm C}_{I_1,\cdots,I_{N}}[\mathcal{W},\mathcal{X}^{(1)},\mathcal{X}^{(2)}, \cdots,\mathcal{X}^{(N)}],
\label{Eq:multilinear-model}
\end{equation}
where $\mathcal{W} \in \mathbb{R}^{I_{1}\times\cdots\times I_{N} \times O}$ is a general $N$-order weight tensor. Note that the number of trainable parameters scales exponentially as $O(\prod_{n = 1}^{N}I_{n})$. To reduce the computation cost, $\mathcal{W}$ can be approximated by a set of low-order tensors which is further contracted according to (\ref{Eq:tensor_train}). The contraction of TT is equivalent to the feed forward process in neural networks. For example, the red arrows in \fref{Fig:TT-structure} indicate a forward direction of TT. The input vector $\mathcal{X}^{(1)}$ is contracted with the weight tensor $\mathcal{W}^{(1)}$ to produce the output tensor $\mathcal{Y}^{(1)} \in \mathbb{R}^{V_{1}}$. The second weight tensor $\mathcal{W}^{(2)}$ and the second input vector $\mathcal{X}^{(2)}$ are contracted to generate an intermediate matrix $\mathcal{A}^{(2)} \in \mathbb{R}^{V_{1}, V_{2}}$, which is then contracted with $\mathcal{Y}^{(1)} \in \mathbb{R}^{V_{1}}$ to generate the second output tensor $\mathcal{Y}^{(2)} \in \mathbb{R}^{V_{2}}$. The final output is denoted by $\mathcal{Y}^{(N)} \in \mathbb{R}^{O}$, whose elements can be written as
\begin{align}
y^{(N)}_{o} = \sum_{v_{1} \cdots v_{N-1}} \sum_{i_{n} \cdots i_{N}} w^{(1)}_{i_{1}v_{1}} x^{(1)}_{i_1} \cdots w^{(N)}_{i_{N}v_{N-1}o} x^{(N)}_{i_N}, \label{Eq:TT}
\end{align}
which clearly encodes the $N$-order interactions among the input features, without low-order terms.

TT can be optimized by gradient descent with respect to the mean squared error. In \cite{stoudenmire2016supervised}, two nearby tensors are contracted to form a new high-order tensor at each step, and then the gradient is calculated by taking the derivative of the cost function with respect to this tensor. After updating the parameters, a singular value decomposition is performed to restore the updated tensor into two low-order tensors. This method sweeps back and forth along the TT and iteratively minimizes the cost function. Notably, at each step all of the tensor nodes must be contracted which may involve hundreds of multiplications, causing the gradient vanishing or exploding problem. Several procedures must be taken to counteract this effect, e.g., normalizing the input and output at each step, dividing the bond tensor with its largest element, etc.  This algorithm has been implemented in the TensorNetwork library \cite{efthymiou2019tensornetwork} which uses TensorFlow as the backend for speeding up the tensor computations.

\begin{figure*}
	\centering
	\includegraphics[width=0.90\textwidth]{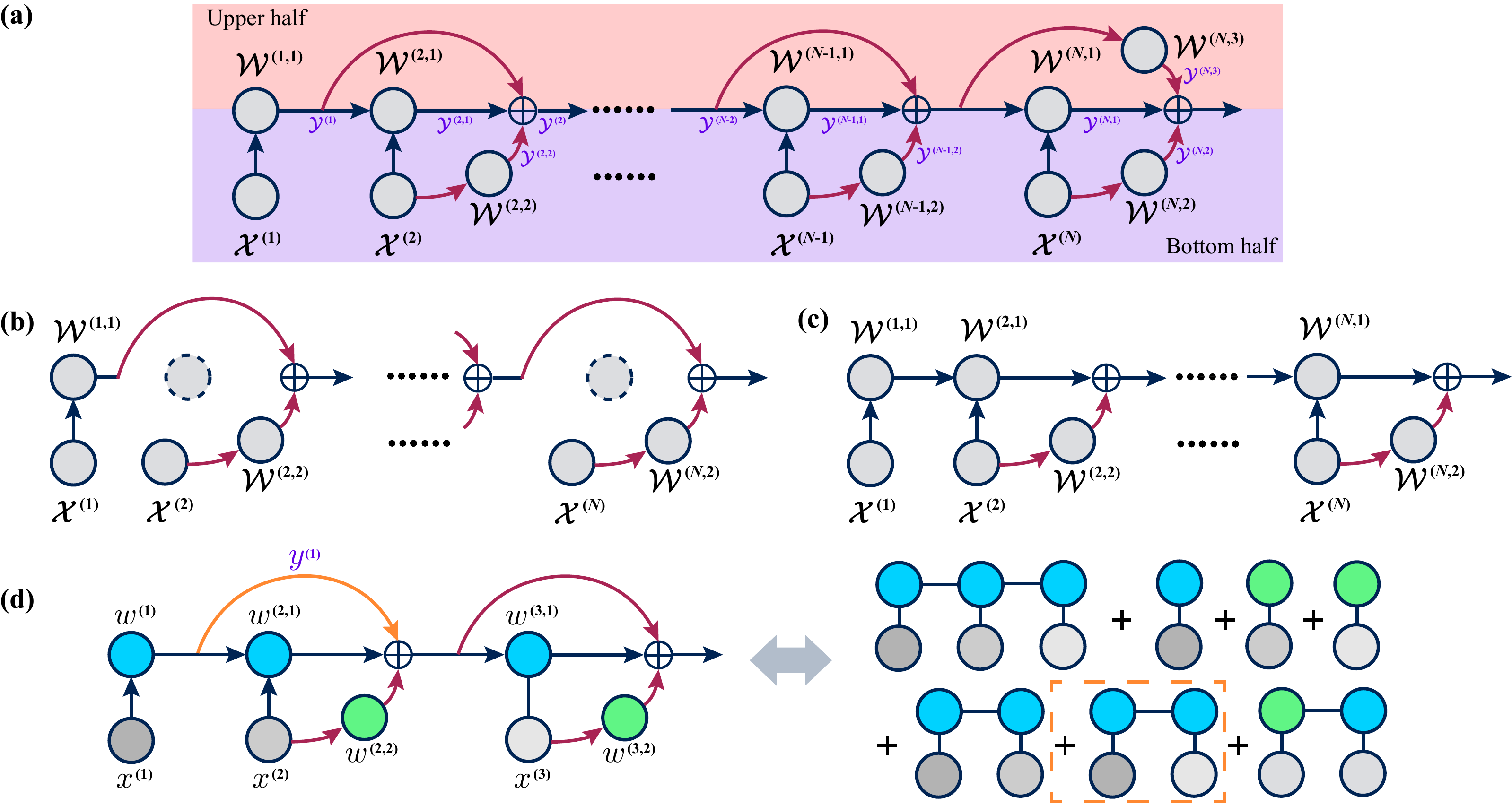}
	\caption{The graphical representation of ResTT. \textbf{(a)} Pipeline of the general ResTT. At each summing junction, two shortcuts (in red line) are added to the output. The arrows indicate the contraction direction. \textbf{(b-c)} The ResTT forms of an $N$-node fully-connected layer and the discrete Volterra series up to the $N$-th order, respectively.   \textbf{(d)} An example of ResTT with three inputs and its equivalent form.
	\vspace{-10pt}}
	\label{Fig:RTT}
\end{figure*}

\section{Residual tensor train}
\label{3}

Residual structure is widely used in the design of deep neural networks. In ResNet, the identity mapping enabled by skip connections allows the input signal to propagate directly from one layer to any other layer, which solves the problem of network degradation to a certain extent \cite{he2016identity}. Besides, the residual network can be regarded as an ensemble model that consists of a series of path sets \cite{veit2016residual}, thereby improving the robustness of the model. Similar to ResNet, the ResTT is proposed to solve the following issues:
\begin{itemize}
	\item The plain TT only models the $N$-order interaction between all the input features while ignores the low-order interactions between a subset of features, such as linear or bilinear correlations. More importantly, the plain TT lacks the flexibility in designing the combination of feature interactions with the required orders. 
	\item The current TT models are difficult to train with a large number of tensor nodes. As residual connection has been proven to be effective in stabilizing the training process for deep neural networks, it can also be added to the TT network to improve its training stability.
\end{itemize}

\subsection{The General Model}
\label{3-1}

The diagrammatic representation of a general ResTT model is shown in \fref{Fig:RTT} (a), in which the contractions are conducted from left to right. Firstly, $\mathcal{X}^{(1)}$ is contracted with $\mathcal{W}^{(1,1)}$ to generate the output $\mathcal{Y}^{(1)} \in \mathbb{R}^{V_{1}}$ by
\begin{equation}
\mathcal{Y}^{(1)} = {\rm C}_{I_1}[\mathcal{X}^{(1)},\mathcal{W}^{(1,1)}].
\label{Eq:x1_contraction}
\end{equation}
Then $\mathcal{X}^{(2)}$ is contracted with $\mathcal{Y}^{1}$ and $\mathcal{W}^{(2,1)}$ to generate the normal output $\mathcal{Y}^{(2,1)} \in \mathbb{R}^{V_{2}}$ by
\begin{equation}
\mathcal{Y}^{(2,1)} = {\rm C}_{V_1,I_2}[\mathcal{Y}^{(1)},\mathcal{W}^{(2,1)},\mathcal{X}^{(2)}].
\label{Eq:x2_contraction}
\end{equation}
The virtual bond dimension of the tensor train is assumed to be the same, i.e., $V_{1} = V_{2} = \cdots = V_{N-1} = r$, and thus we can directly add the skip connection $\mathcal{Y}^{(1)}$ to the output $\mathcal{Y}^{(2,1)}$. Besides, $\mathcal{X}^{(2)}$ is contracted with an additional node $\mathcal{W}^{(2,2)} \in \mathbb{R}^{I_{2} \times V_{2}}$ to generate the low-order interaction terms that include the features in the input $\mathcal{X}^{(2)}$, according to the following formula
\begin{equation}
\mathcal{Y}^{(2,2)} = {\rm C}_{I_2}[\mathcal{X}^{(2)},\mathcal{W}^{(2,2)}].
\label{Eq:linear_map}
\end{equation}
The output of the second layer of ResTT is given by
\begin{equation}
\mathcal{Y}^{(2)} = \mathcal{Y}^{(2,1)} + \mathcal{Y}^{(1)} +  \mathcal{Y}^{(2,2)}.
\label{Eq:y2_new}
\end{equation}
Note that the skip connection $\mathcal{Y}^{(1)}$ produces the low-order terms that include features in the input $\mathcal{X}^{(1)}$. By repeating this process we obtain
\begin{align}
&\mathcal{Y}^{(n,1)} = {\rm C}_{V_{n-1},I_n}[\mathcal{Y}^{(n-1)},\mathcal{W}^{(n,1)},\mathcal{X}^{(n)}], \nonumber\\
&\mathcal{Y}^{(n,2)} = {\rm C}_{I_n}[\mathcal{X}^{(n)},\mathcal{W}^{(n,2)}], \nonumber\\
&\mathcal{Y}^{(n)} = \mathcal{Y}^{(n,1)} + \mathcal{Y}^{(n-1)} + \mathcal{Y}^{(n,2)}.
\label{Eq:yn_new}
\end{align}
For the last node a linear mapping is defined by
\begin{equation}
\mathcal{Y}^{(N,3)} = {\rm C}_{V_{N-1}}[\mathcal{W}^{(N,3)},\mathcal{Y}^{(N-1)}]
\end{equation}
with $\mathcal{W}^{(N,3)} \in \mathbb{R}^{V_{N-1} \times O}$, and the final output of ResTT can be written as
\begin{equation}
\mathcal{Y}^{(N)} = \mathcal{Y}^{(N,1)} + \mathcal{Y}^{(N,3)} + \mathcal{Y}^{(N,2)}.
\label{Eq:Final output}
\end{equation}
The detailed expression of $\mathcal{Y}^{(N)}$ is given in Appendix \ref{Ap.1}. From the detailed expression, we can see that the output of a general ResTT contains the interaction terms from $1$-order (linear) to $N$-order ($N$-linear).

\subsection{Special Cases}
\label{3-2}

It is clear from (\ref{Eq:yn_new}) that different connections at each layer generate interaction terms with different orders of features interactions. Therefore, it is convenient to modify the combination of multilinear terms contained in the final output by adding or deleting the specific connections between the tensor nodes. In particular, many famous models are found to be special cases of the general ResTT. For example, if only the residual connections shown in \fref{Fig:RTT} (b) are kept, ResTT will degrade to an $N$-node fully-connected layer as
\begin{equation}
\mathcal{Y}^{(N)} = \sum_{n=1}^{N}{\rm C}_{I_n}[\mathcal{X}^{(n)},\mathcal{W}^{(n,2)}].
\label{Eq:Final output_FN}
\end{equation}
In this case, the model captures only the linear correlations.

Meanwhile, if we assume the input and output are scalars, i.e., $\mathcal{Y}^{(N)} \in \mathbb{R}$, the general ResTT will degrade to the famous discrete Volterra series \cite{s1985fading} when the residual connections in the upper half of \fref{Fig:RTT} (a) are deleted (See \fref{Fig:RTT} (c)). The resulting Volterra series are given by \begin{eqnarray}
&y&=\sum_{i_{N}} h^{(N)}_{i_{N}} x^{(N)}_{i_N} + \sum_{i_{N}i_{N-1}} h^{(N-1)}_{i_{N} i_{N-1}} x^{(N)}_{i_N}x^{(N-1)}_{i_{N-1}} + \cdots \nonumber\\
&&+\sum_{i_{1}, \cdots, i_{N}} h^{(1)}_{i_{1} \ldots i_{N}} \prod_{j=1}^{N} x^{(j)}_{i_j}
\end{eqnarray}
with
\begin{eqnarray}
&h^{(N)}_{i_{N}} = w^{(N,2)}_{i_{N}},& \nonumber\\
&h^{(N-1)}_{i_{N-1} i_{N}} =   \sum_{v_{N-1}}w^{(N-1,2)}_{i_{N-1}v_{N-1}} w^{(N,1)}_{v_{N-1}i_{N}},&\nonumber\\
& \cdots &\nonumber\\
&h^{(1)}_{i_{1} \ldots i_{N}} =   \sum_{v_1,\cdots,v_{N-1}}w^{(1,1)}_{i_{1}v_{1}} w^{(2,1)}_{v_{1}i_{2}v_{2}} \cdots w^{(N,1)}_{i_{N}v_{N-1}}&
\end{eqnarray}
being the tensorization form of the elements of Volterra kernels denoted by $\mathcal{H}^{(N)}\ \in \mathbb{R}^{I_{N}}, \mathcal{H}^{(N-1)}\ \in \mathbb{R}^{I_{N}, I_{N-1}},...,\mathcal{H}^{(1)} \in \mathbb{R}^{I_{1}, I_{2},\cdots,I_{N}}$, where the $k$-order tensor $\mathcal{H}^{(k)}$ is called the $k$-th Volterra kernel. 

In the last, we provide a simple example to further illustrate the flexibility of ResTT. As shown in \fref{Fig:RTT} (d), the final output of the ResTT can be written as
\begin{align}
y & = w^{(1)}x^{(1)} + w^{(2,2)}x^{(2)} + w^{(3,2)}x^{(3)} +\nonumber \\
&w^{(1)}w^{(2,1)}x^{(1)}x^{(2)}  + w^{(1)}w^{(3,1)}x^{(1)}x^{(3)} + \nonumber\\
&w^{(2,2)}w^{(3,1)}x^{(2)}x^{(3)} + w^{(1)}w^{(2,1)}w^{(3,1)}x^{(1)}x^{(2)}x^{(3)}.
\label{Eq.3-dimen-case-RTT1}
\end{align}
Suppose the  skip connection of $y^{(1)}$ (the orange line in \fref{Fig:RTT} (d)) has been removed. Then the output of ResTT will change to
\begin{align}
y & = w^{(1)}x^{(1)} + w^{(2,2)}x^{(2)} + w^{(3,2)}x^{(3)} +\nonumber \\
&w^{(1)}w^{(2,1)}x^{(1)}x^{(2)} + w^{(2,2)}w^{(3,1)}x^{(2)}x^{(3)}  \nonumber\\
& w^{(1)}w^{(2,1)}w^{(3,1)}x^{(1)}x^{(2)}x^{(3)},
\label{Eq.3-dimen-case-RTT2}
\end{align}
in which the 2-order interaction between the input features $x^{(1)}$ and $x^{(3)}$ is discarded.

\subsection{Optimization}
\label{3-3}

Since the operation of tensor contraction is similar to the linear transformation layer of conventional neural networks, we can adopt the standard backpropagation algorithm \cite{lecun1989backpropagation} to optimize the ResTT. Here we briefly show the gradient calculation of the weight tensor $\mathcal{W}^{(k,1)}$ as follows
\begin{align}
\Delta \mathcal{W}^{(k,1)} =&-\dfrac{\partial c}{\partial\mathcal{W}^{(k,1)}}\nonumber\\
=&-\dfrac{\partial c}{\partial \mathcal{Y}^{(N)}}\cdot\dfrac{\partial \mathcal{Y}^{(N)}}{\partial \mathcal{Y}^{(N-1)}}\cdot\dfrac{\partial \mathcal{Y}^{(N-1)}}{\partial \mathcal{Y}^{(N-2)}} \cdots \dfrac{\partial \mathcal{Y}^{(k)}}{\partial \mathcal{W}^{(k,1)}}\nonumber\\
=&-\dfrac{\partial c}{\partial \mathcal{Y}^{(N)}}\cdot(\mathcal{I} + \mathcal{A}^{(N)}) \cdots \dfrac{\partial \mathcal{Y}^{(k)}}{\partial \mathcal{W}^{(k,1)}}\nonumber\\
=&-\dfrac{\partial c}{\partial \mathcal{Y}^{(N)}}\cdot(\mathcal{I} + \mathcal{A}^{(N)}) \cdots (\mathcal{I} + \mathcal{A}^{(k+1)})\nonumber\\
&\otimes \mathcal{X}^{(k)} \otimes \mathcal{Y}^{(k-1)},
\label{gradient}
\end{align}
with the intermediate matrix $\mathcal{A}^{(k)}$ whose elements are given by
\begin{equation}
a^{(k)}_{v_{k}v_{k-1}} = \sum_{i_{k}}w^{(k,1)}_{v_{k}i_{k}v_{k-1}}x_{i_{k}}.
\label{E_Ak1_ResTT}
\end{equation}
Here $\mathcal{I}$ is an $r \times r$ identity matrix and $c$ is the current value of the cost function. Dot notation $\cdot$ in \eref{gradient} means that the derivatives are connected by matrix multiplication along the compatible indices. Similarly, the gradient of the weight tensor $\mathcal{W}^{(k,2)}$ is given by
\begin{align}
\Delta \mathcal{W}^{(k,2)} =&-\dfrac{\partial c}{\partial\mathcal{W}^{(k,2)}}\nonumber\\
=&-\dfrac{\partial c}{\partial \mathcal{Y}^{(N)}}\cdot\dfrac{\partial \mathcal{Y}^{(N)}}{\partial \mathcal{Y}^{(N-1)}}\cdot\dfrac{\partial \mathcal{Y}^{(N-1)}}{\partial \mathcal{Y}^{(N-2)}} \cdots \dfrac{\partial \mathcal{Y}^{(k)}}{\partial \mathcal{W}^{(k,2)}}\nonumber\\
=&-\dfrac{\partial c}{\partial \mathcal{Y}^{(n)}} \cdot(\mathcal{I} + \mathcal{A}^{(N)}) \cdots(\mathcal{I} + \mathcal{A}^{(k)}) \otimes \mathcal{X}^{(k)} .
\label{gradient_2}
\end{align}

The details of the ResTT implementation for supervised learning are summarized in \Alref{Al:implementation}.

\begin{figure}[!t]
	\renewcommand{\algorithmicrequire}{\textbf{Input:}}
	\renewcommand{\algorithmicensure}{\textbf{Initialize:}}
	\removelatexerror
    		\begin{algorithm}[H]
			\caption{ResTT for Supervised Learning}
			\label{Al:implementation}
			\begin{algorithmic}[1]
				\REQUIRE Features $ \mathcal{X}^{(1)} \in \mathbb{R}^{I_{1}},  \mathcal{X}^{(2)} \in \mathbb{R}^{I_{2}},\cdots, \mathcal{X}^{(N)}\in \mathbb{R}^{I_{N}}$; \\
				\ \quad  Label $ \mathcal{Y} \in \mathbb{R}^{O}$;
				\ENSURE The number of epochs $K$; \\
				\qquad \quad Weight tensors  $ \mathcal{W}^{(1)} \in \mathbb{R}^{I_{1} \times V_{1}}, $\\
				\qquad \quad$\mathcal{W}^{(2,1)} \in \mathbb{R}^{V_{1}\times I_{2}\times V_{2}}, \mathcal{W}^{(2,2)} \in \mathbb{R}^{I_{2}\times V_{2}},$\\
                \qquad \quad$\cdots,$\\
				\qquad \quad $\mathcal{W}^{(N,1)}\in \mathbb{R}^{V_{N-1}\times I_{N}\times O}, \mathcal{W}^{(N,2)}\in \mathbb{R}^{I_{N}\times O},$\\
				\qquad \quad $ \mathcal{W}^{(N,3)}\in \mathbb{R}^{V_{N-1}\times O}$;\\
				\FOR{$k = 1, \cdots, K$}
				\STATE $\mathcal{Y}^{(1)} = {\rm C}_{I_{1}}[\mathcal{X}^{(1)}, \mathcal{W}^{(1)}]$.
				\FOR{$n=2, \cdots, N$}
				\STATE $\mathcal{Y}^{(n,1)} ={\rm C}_{V_{n-1},I_n}[\mathcal{Y}^{(n-1)},\mathcal{W}^{(n,1)},\mathcal{X}^{(n)}]$.
				\STATE $\mathcal{Y}^{(n,2)} = {\rm C}_{I_n}[\mathcal{X}^{(n)},\mathcal{W}^{(n,2)}]$.
				\IF{$n \neq N$}
				\STATE $\mathcal{Y}^{(n)} = \mathcal{Y}^{(n,1)} + \mathcal{Y}^{(n-1)} + \mathcal{Y}^{(n,2)}$.
				\ENDIF
				\ENDFOR
				\STATE $\mathcal{Y}^{(N,3)} = {\rm C}_{V_{N-1}}[\mathcal{W}^{(N,3)},\mathcal{Y}^{(N-1)}]$.
				\STATE $\mathcal{Y}^{(N)} = \mathcal{Y}^{(N,1)} + \mathcal{Y}^{(N,2)} + \mathcal{Y}^{(N,3)}$.
				\STATE Calculate the cost function: $c(\mathcal{Y}, \mathcal{Y}^{(N)})$.
				\STATE Update the weight tensors by backpropagation.
				\ENDFOR
			\end{algorithmic}	
		\end{algorithm}
\end{figure}

\vspace{-10pt}
\subsection{Complexity Analysis}
\label{3-4}
Here we compare the memory complexity and time complexity of three multilinear models, including the general tensorized model in \eref{Eq:multilinear-model}, the plain TT in \eref{Eq:tensor_train} and ResTT. The memory complexity of \eref{Eq:multilinear-model} is $O(\prod_{n = 1}^{N}i_{k})$. In contrast, the memory complexities of TT and ResTT are $O(\sum_{n = 1}^{N}I_{n}r^2)$ and $O(\sum_{n = 1}^{N}(I_{n}r^2 + I_{n}r))$, respectively. The time complexities for TT and ResTT are $O(\sum_{n = 1}^{N}I_{n}r^3)$ and $O(\sum_{n = 1}^{N}(I_{n}r^3 + I_{n}r))$, respectively, while \eref{Eq:multilinear-model} has a complexity of $O(\prod_{n = 1}^{N}I_{k})$.

\section{Mean-field Analysis}
\label{4}
Mean-field analysis \cite{saxe2013exact, schoenholz2016deep} provides an efficient way to assess the training stability of a deep neural network. In this paper we extend this analysis to study the characteristics of signal propagations in TT and ResTT. First we assume that all the weights in TT and ResTT are initialized from the same Gaussian distribution $\mathcal N(0,\sigma^{2}_{w}/r)$, and then calculate the evolution of the variances of the outputs $\{\mathcal{Y}^{(k)}\}$ and the average magnitude of the gradients. The detailed derivations are given in Appendix \ref{Ap.2}.


\subsection{Tensor Train}
\label{4-1}
As all the weights are \textit{i.i.d} with zero mean, the output $\mathcal{Y}^{(k)}$ satisfies
\begin{equation}
{\mathbb E}(y^{(k)}_{v_{k}}) = 0,
\end{equation}
 and the variance of the output, denoted by $q^{(k)}$, is calculated as
\begin{equation}
q^{(k)} ={\mathbb E}(\sum_{i_{k}}(x^{(k)}_{i_{k}})^2)\sigma^{2}_{w}q^{(k-1)} = s^{(k)} q^{(k-1)},
\label{Eq:q_TT}
\end{equation}
with $s^{(k)} = {\mathbb E}(\sum_{i_{k}}(x^{(k)}_{i_{k}})^2)\sigma^{2}_{w}$. Hence, the relation between the input- and output- variances at each layer is characterized by the slope factor $s^{(k)}$. As can be seen from \eref{Eq:q_TT}, the slope factor is influenced by the variance of the input data, which is significantly different from the conventional deep learning model where the evolution of the output at each layer can only be influenced by the variance of the weights. In addition, the backward propagation in TT is studied by calculating the input-output Jacobian, which relates the gradients to the weight tensor at a given position. The Jacobian for the input $\mathcal{Y}^{(k-1)}$ is denoted by $\mathcal{J}^{(k)} \in \mathbb{R}^{V_{k-1} \times I_{k} \times V_{k}}$ and calculated as
\begin{equation}
\mathcal{J}^{(k)} = \dfrac{\partial \mathcal{Y}^{(N)}}{\partial \mathcal{Y}^{(N-1)}}\cdots \dfrac{\partial \mathcal{Y}^{(k)}}{\partial \mathcal{Y}^{(k-1)}}
= \mathcal{A}^{(N)} \cdots  \mathcal{A}^{(k+1)} \otimes \mathcal{X}^{(k)}\nonumber\\
\label{Eq:J_TT}
\end{equation}
with the intermediate matrix $\mathcal{A}^{(k)}$ defined in \eref{E_Ak1_ResTT}. The quadratic mean $\chi^{(k)}$ of the elements in the Jacobian characterizes the magnitude of the gradients in backpropagation, which is given by
\begin{equation}
\chi^{(k)} = s^{(k)} \chi^{(k+1)}.
\label{Eq:chi_TT}
\end{equation}
According to \eref{Eq:q_TT} and \eref{Eq:chi_TT}, the signals in the forward- and backward- propagations will explode in a long TT when $s^{(k)}>1$. In contrast, if $s^{(k)}<1$, the signals will tend to vanish. Therefore, the condition of stability for the training is given by the following formula
\begin{equation}
s^{(k)}=1, \quad k\geq1.
\label{Eq:sk-condi}
\end{equation}
Since the $s^{(k)}$ is determined by the input data, we have to preprocess the input in order to stabilize the training. For example, by employing the trigonometric mapping \cite{vapnik2013nature} to the input and transforming the feature vector $\mathcal{X}^{(k)}$ as
\begin{equation}
{\phi}(\mathcal{X}^{(k)}) = \frac{1}{\sqrt{I_k}}[\cos (\frac{\pi }{2}{\mathcal{X}^{(k)}}),\sin (\frac{\pi }{2}{\mathcal{X}^{(k)}})],
\label{map-TT}
\end{equation}
we have ${\mathbb E}(\sum_{i_{k}}(x^{(k)}_{i_{k}})^2)=1$. Then \eref{Eq:sk-condi} can be satisfied by letting $\sigma^{2}_{w}=1$. In this case we get
\begin{equation}
q^{(k)}=q^{(k-1)},\quad \chi^{(k)}=\chi^{(k+1)},
\label{Eq:AppoInia}
\end{equation}
which is the critical condition for the stable training of TT.

\subsection{Residual Tensor Train}
\label{4-2}
In ResTT, the $k$-th output vector also satisfies ${\mathbb E}(\mathcal{Y}^{(k)}_{v_{k}})=0$, and the variance of this vector is calculated as
\begin{equation}
q^{(k)} =(s^{(k)}+1)q^{(k-1)}+ s^{(k)}.
\label{Eq:q_ResTT}
\end{equation}
In backward propagation, the quadratic mean of the input-output Jacobian for the input $\mathcal{Y}^{(k-1)}$ is calculated as
\begin{equation}
\chi^{(k)} = (s^{(k)}+1)\chi^{(k+1)},
\label{Eq:J_ResTT_1}
\end{equation}
which shares the same slope factor as the forward propagation. Hence, $(s^{(k)}+1)$ determines the evolution of the signals in the training of ResTT. Note that for any choice of $\sigma^{2}_{w}$, $(s^{(k)}+1)$ is always larger than 1, which means the signals in the forward and backward propagation will never vanish. Meanwhile, in order to prevent the signals from exploding, the following condition has to be satisfied
\begin{equation}
s^{(k)} \ll 1.
\label{Eq:C_ResTT}
\end{equation}
According to \eref{Eq:q_TT}, as long as $\sigma^{2}_{w} \ll 1$, the condition of stability \eref{Eq:C_ResTT} will be satisfied given that the variance of the input data is bounded. Compared to TT which requires ${\mathbb E}(\sum_{i_{k}}(x^{(k)}_{i_{k}})^2)\sigma^{2}_{w}=1$, the condition $\sigma^{2}_{w} \ll 1$ is significantly relaxed.


\section{Experiments}
\label{5}
We have conducted comprehensive experiments on the synthetic dataset, two image classification datasets and two practical examples with limited data. In terms of experimental setup, we optimize the models by Adam optimizer \cite{kingma2014adam}. The learning rate is chosen from $\{1e^{-2},1e^{-3},1e^{-4}\}$. All models are trained for 100 epochs. We fix the mini-batch size as $512$ and set the weight decay as $1e^{-6}$ for all tasks. The implementations are based on Pytorch \cite{paszke2019pytorch} and the models are trained on a single NVIDIA GTX 1080Ti.

\begin{figure}
	\centering
	\includegraphics[width=0.42\textwidth]{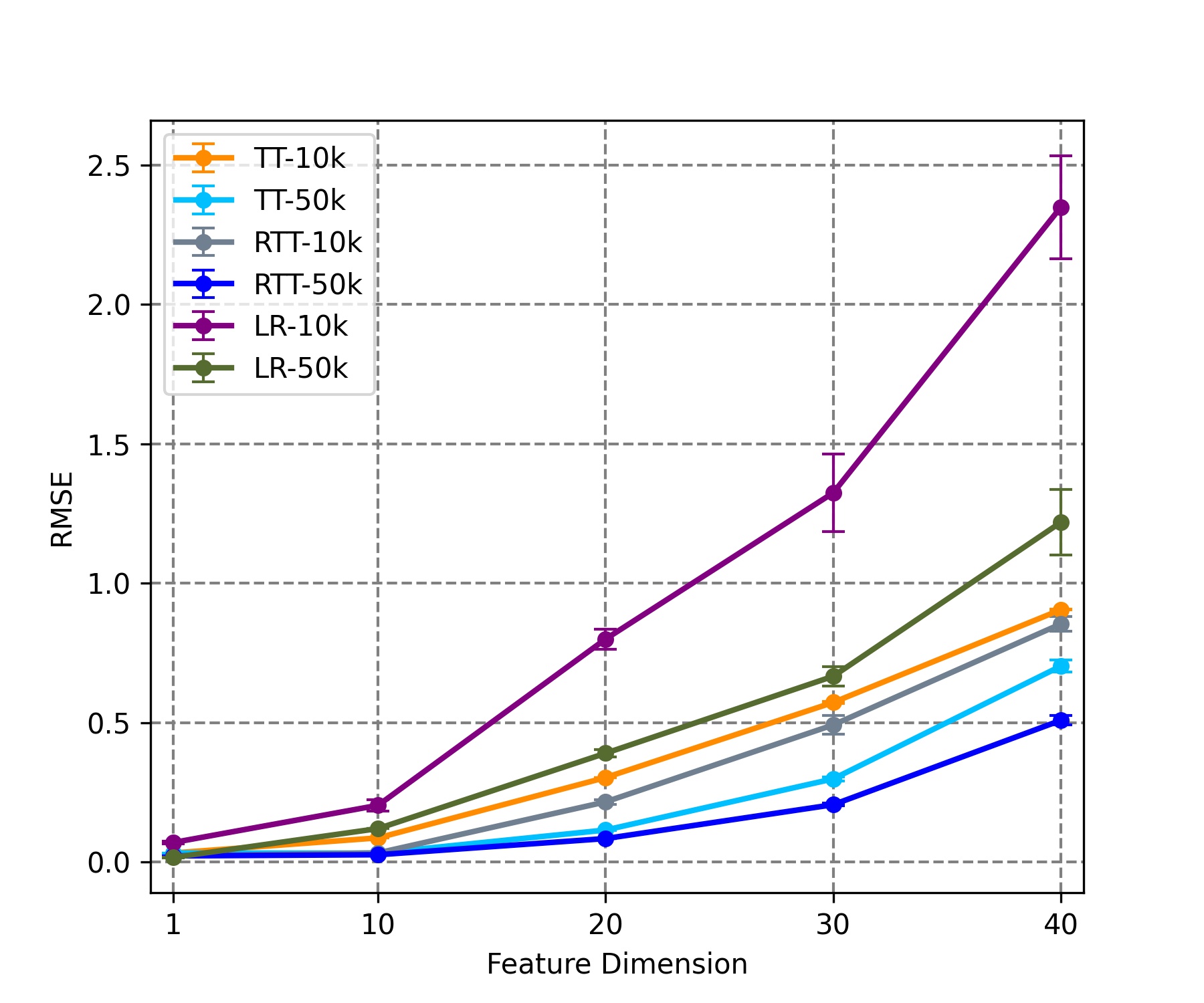}
	\caption{Comparisons between ResTT, TT and LR on synthetic datasets with different data sizes. The Y-axis is RMSE and X-axis is the dimension of feature. RTT-$N$ is the notation for the ResTT model that trains on $N$ samples.}
	\label{Fig:sy_result}
\end{figure}

\subsection{Synthetic dataset}
\label{5-1}
In order to compare the capability of ResTT in modelling multiple multilinear feature correlations, we generate a synthetic dataset by
\begin{align}
\hat{y} = \sum_{i=1}^{d} w^{(1)}_{i} x^{(1)}_{i} + \sum_{i,j=1}^{d} w^{(2)}_{ij} x^{(1)}_{i} x^{(2)}_{j} \nonumber \\
 + \sum_{i,j,k=1}^{d} w^{(3)}_{ijk} x^{(1)}_{i} x^{(2)}_{j}x^{(3)}_{k},
 \label{Eq:synthetic}
\end{align}
where the elements of $\mathcal{W}^{(1)} \in \mathbb{R}^{d},\mathcal{W}^{(2)} \in \mathbb{R}^{d \times d},\mathcal{W}^{(3)} \in \mathbb{R}^{d \times d \times d} $ are randomly sampled from $\mathcal N (0,0.1)$, and $\mathcal{X}^{(1)},\mathcal{X}^{(2)},\mathcal{X}^{(3)} \in \mathbb{R}^{d}$ are randomly sampled from $\mathcal N (0,0.5)$. The training sizes are $10000$ and $50000$, while the test size is $10000$ for all cases. The feature dimension $d$ is selected from $\{1, 10, 20 ,30 ,40\}$. We conduct experiments on TT in \eref{Eq:TT} with $r=20$, ResTT with $r=20$ and the Linear Regression (LR) model. Each model has been run 5 times to average the effect of random initialization.

We use the Root Mean Square Error (RMSE) as the evaluation metric. The results are shown in \fref{Fig:sy_result}. It is clear that TT performs better than LR, with its ability to model high-order feature interaction instead of the low-order ones. Meanwhile, the performance of TT is worse than that of ResTT, since ResTT can model a comprehensive set of feature interactions. \fref{Fig:sy_result} also shows that the advantage of ResTT over TT and LR becomes more dominant with the increase of feature dimension $d$. 

\subsection{Image classification}
\label{5-2}
In line with the previous works \cite{stoudenmire2016supervised,efthymiou2019tensornetwork,glasser2018supervised}, we firstly use the MNIST and Fashion-MNIST datasets to evaluate the classification performance of ResTT. However, it should be noted that the intended application of the current ResTT is not for large-scale image datasets. On one hand, images have to be flattened into a list as the input to the current ResTT, which makes it difficult for ResTT to outperform convolutional neural networks (CNN) that take advantage of spatial correlations. On the other hand, the current ResTT is an extended linear model without using any advanced network components such as nonlinear activation, which makes it a non-fair competition between ResTT and other deep learning models on large-scale datasets.
\subsubsection{Datasets}
\label{5-2-1}
\begin{itemize}
	\item [a)] \textit{MNIST:} The MNIST is a dataset of handwritten digits from 0 to 9, which contains $60,000$ training samples and $10,000$ testing samples with $28 \times 28$ gray-scale pixels. Each pixel takes integer in the range $[0,255]$. Following \cite{stoudenmire2016supervised}, each pixel value of the image is normalized to $[0,1]$ and mapped into a two-dimensional space by a trigonometric function defined in \eref{map-TT}. Hence, each feature of this task is a 2-dimensional vector. Each image is scaled down to $14 \times 14$ by averaging the clusters of four adjacent pixels and then flattened into a list to adapt to the structure of TT.
	
	\item [b)] \textit{Fashion-MNIST:} The Fashion-MNIST is a dataset of Zalando's article images, which consists of a training set of $60,000$ samples and a test set of $10,000$ samples. Each image is a $28 \times 28$ gray-scale image, associated with a label from 10 classes. Similar to MNIST, each pixel value is normalized to $[0,1]$ and mapped to a two-dimensional space by the trigonometric function. Again, each image is scaled down to $14 \times 14$ by averaging the clusters of four adjacent pixels and then flattened into a list to adapt to the structure of TT.

\end{itemize}

\subsubsection{Models}
\label{5-2-2}
The following models based on tensor network are tested in the experiments:
\begin{itemize}
	\item [a)] \textit{MPS \cite{stoudenmire2016supervised}:} Matrix Product State (MPS) is a special TT model in which all the tensor nodes are kept unitary and trained by the DMRG-like optimization algorithm.
	
	\item [b)] \textit{TTN \cite{liu2018machine}:} Tree tensor network is a 2-layer tensor network model that supports hierarchical feature extraction, which uses a training algorithm derived from the multipartite entanglement renormalization ansatz.
	
	\item [c)] \textit{EPS-SBS \cite{glasser2018supervised}:} Entangled Plaquette State (EPS) is defined as a product of tensors on overlapping clusters of variables and String-Bond State (SBS) is defined by placing MPS over strings on a graph which needs not to be a one-dimensional lattice. EPS-SBS is a tensor network that consists of an EPS followed by a SBS and optimized by stochastic gradient descent.
		
	\item [d)] \textit{Snake-SBS \cite{glasser2018supervised}:} Snake-SBS consists of four overlapping SBS structures in a snake pattern and is optimized by stochastic gradient descent.
	
	\item [e)] \textit{TT-BN:} Tensor train with Batch Normalization (BN). In this model, a BN layer \cite{ioffe2015batch} is added after each contraction. TT-BN is trained by backpropagation.
	
	\item [f)] \textit{MeanTT:} Mean-field Tensor Train is the vanilla TT model which is initialized using the stability condition derived in \sref{4-1}. MeanTT is trained by backpropagation.
\end{itemize}

\begin{table}[]
 \renewcommand\arraystretch{1.3}
 \centering
 \caption{The number of trainable parameters for TT and ResTT with different virtual bond dimensions $r$ on the MNIST and Fashion-MNIST datasets.}
 \label{Tab:Params}
 \begin{tabular}{c|c|c}
  \hline
  $r$    & TT   & ResTT \\ \hline
  10 & 38840 &42740 \\ \hline
  20 &155280 &163080\\ \hline
  30 &349320 &361020\\ \hline
  40 &620960 &636560\\ \hline
  50 &970200 &989700\\ \hline
  100 &3880400 &3919400\\
  \hline
 \end{tabular}
\end{table}

\begin{figure*}[!htp]
	\centering
	\includegraphics[width=0.9\textwidth]{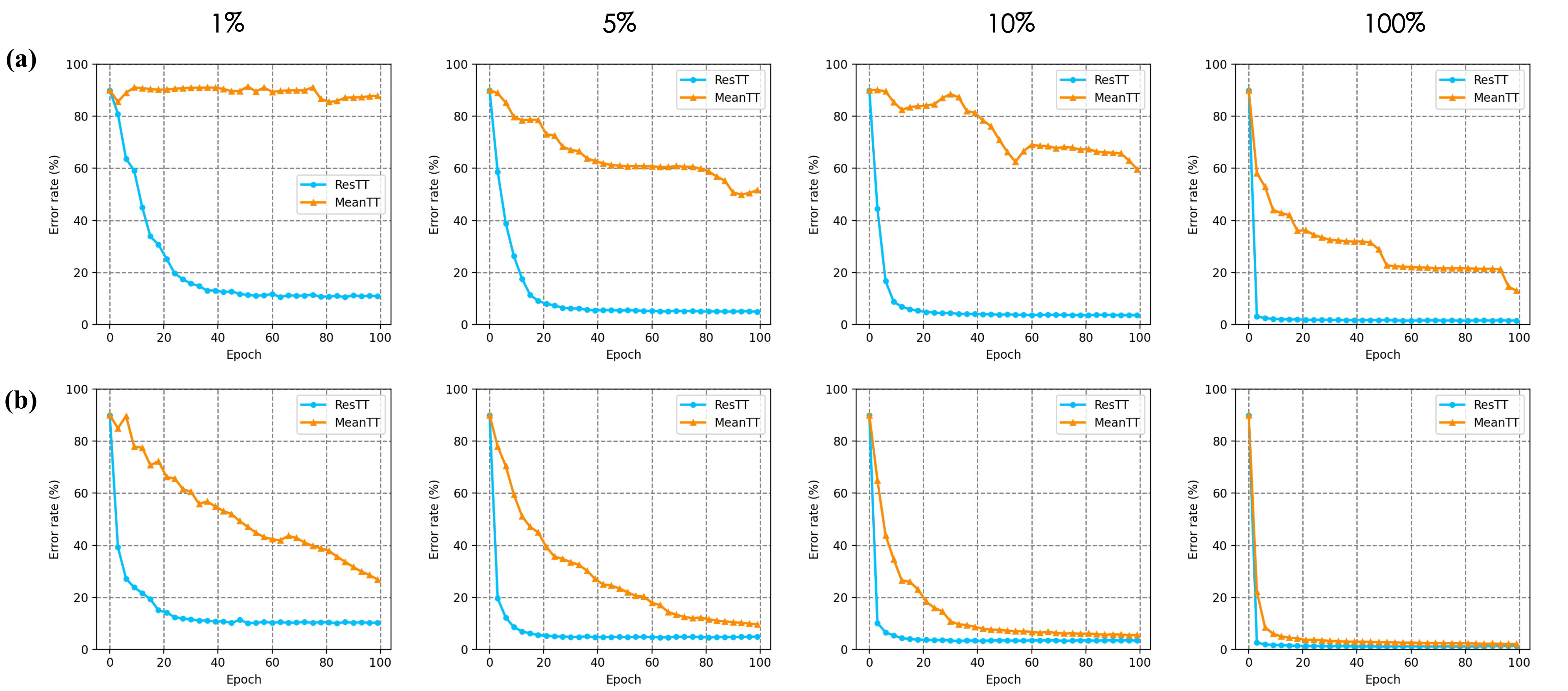}
	\caption{Test accuracy of ResTT (blue curve) and MeanTT (orange curve) versus the training epochs on MNIST. \textbf{(a)}, \textbf{(b)} are the model performances with $r = 20$ and $r = 100$, respectively.}
	\label{Fig:Train}
\end{figure*}

\begin{table}[]
	\renewcommand\arraystretch{1.3}
	\centering
	\caption{Comparisons of the classification accuracy ($\%$) of ResTT and MeanTT with the state-of-the-art tensor network models on the test sets of MNIST and Fashion-MNIST.}
	\label{Tab:Cresult}
	\begin{tabular}{c|c|c}
		\hline
		Model    & MNIST   & Fashion-MNIST \\ \hline
		TTN\cite{liu2018machine}       & 95\%    & --     \\
		MPS\cite{efthymiou2019tensornetwork}       & 98\%    & 88\%    \\
		EPS-SBS\cite{glasser2018supervised}   & 98.7\%  & 88.6\%  \\
		Snake-SBS\cite{glasser2018supervised} & 99\%    & 89.2\%  \\
		TT-B($r=100$)        & 95.05\% & 85.68\% \\
		MeanTT($r=100$)       & 98.75\% & 88.86\% \\
		ResTT($r=100$)       & \textbf{99.04\%} & \textbf{89.95\%} \\
		\hline
	\end{tabular}
\end{table}

\begin{table}[]
	\renewcommand\arraystretch{1.3}
	\centering
	\caption{The inference time (IT), backpropagation time (BT) and convergence time (CT) (for achieving a test accuracy that is above $98\%$) of ResTT with different virtual bond dimensions $r$ on MNIST dataset.}
	\label{Tab:Time}
	\begin{tabular}{c|c|c|c}
		\hline
		$r$ & IT (s)&BT (s)& CT (s)\\ \hline
		20 &  0.326&0.256&429.59\\ \hline
		30 & 0.333&0.266&256.27 \\ \hline
		40 & 0.338&0.272&255.12\\ \hline
		50 & 0.359 & 0.296 &226.49\\ \hline
		100&0.541 & 0.466& 324.03\\
		\hline
	\end{tabular}
\end{table}

\begin{table*}[]
	\renewcommand\arraystretch{1.5}
	\centering
	\caption{Test accuracy (\%) w.r.t. the fraction of the training samples on MNIST and Fashion-MNIST datasets.}
	\label{Tab:Mresult}
	\scalebox{0.9}{\begin{tabular}{|c|c|c|c|c|c|c|c|c|c|}
			\hline
			Dataset & Method & $r$   & Parm. &Training 1\% & Training 5\% & Training 10\% & Training 20\% & Training 50\% & Training 100\% \\ \hline
			\multirow{4}{*}{MNIST}& MeanTT    & 20  & 0.15M & 30.16$\pm$14.59  & 37.03$\pm$14.95  & 48.53$\pm$26.75   & 68.14$\pm$20.70   & 57.01$\pm$33.17   & 38.54$\pm$33.14    \\ \cline{2-10}
			& MeanTT    & 100 & 3.88M& 85.87$\pm$0.6054 & 94.57$\pm$0.1231 & 96.20$\pm$0.1762  & 97.27$\pm$0.0721  & 98.36$\pm$0.0412  & 98.70$\pm$0.0474   \\ \cline{2-10}
			& ResTT    & 20  & 0.16M& 89.52$\pm$0.2917 & 95.19$\pm$0.1088 & 96.49$\pm$0.0889  & 97.41$\pm$0.0902  & 98.34$\pm$0.0306  & 98.53$\pm$0.0397   \\ \cline{2-10}
			& ResTT    & 100 & 3.92M& \textbf{89.82$\pm$0.2755} & \textbf{95.46$\pm$0.0768} & \textbf{96.74$\pm$0.0706}  & \textbf{97.68$\pm$0.0687}  & \textbf{98.66$\pm$0.0329}  & \textbf{98.96$\pm$0.0304}   \\ \hline
			\hline
			\multirow{4}{*}{Fashion-MNIST}& MeanTT    & 20  & 0.15M& 22.77$\pm$1.071  & 30.72$\pm$0.5816  & 53.11$\pm$0.1682  & 44.67$\pm$0.1979  & 56.29$\pm$0.0804 & 37.44$\pm$0.1035\\ \cline{2-10}
			& MeanTT    & 100 & 3.88M& 77.07$\pm$0.4257 & 82.72$\pm$0.2014 & 84.39$\pm$0.1861 & 85.92$\pm$0.1282  & 87.66$\pm$0.0986  & 88.79$\pm$0.0841  \\ \cline{2-10}
			& ResTT    & 20  & 0.16M& 75.48$\pm$0.4455 & 82.11$\pm$0.2052 & 83.95$\pm$0.0925 & 85.45$\pm$0.0731  & 86.95$\pm$0.0695  & 87.95$\pm$0.0913   \\ \cline{2-10}
			& ResTT    & 100 & 3.92M& \textbf{78.86$\pm$0.3120} & \textbf{84.18$\pm$0.1554} & \textbf{85.69$\pm$0.1116} & \textbf{87.16$\pm$0.1363}  & \textbf{88.93$\pm$0.1132}  & \textbf{89.85$\pm$0.1064}   \\ \hline
	\end{tabular}}
\end{table*}


\subsubsection{Results}
\label{5-2-3}	
Table \ref{Tab:Params} shows that the additional connections in the ResTT do not introduce a significant amount of parameters, which means TT and ResTT can be compared with the same virtual bond dimension $r$. This is because most of the parameters are contained in the high-order tensors generated by $\{\mathcal W^{(n,1)}\}$, while the additional parameters are contained in $\{\mathcal W^{(n,2)}\}$ which only generate low-order tensors. The test accuracy on MNIST and Fashion-MNIST datasets are presented in Table \ref{Tab:Cresult}. It can be observed that ResTT outperforms all the state-of-the-art tensor network models on both MNIST and Fashion-MNIST datasets. In particular, MeanTT achieves better test accuracy than the regularized MPS and TT-BN, which indicates that proper initialization of the weights can lead to improved performance without any regularization techniques. The inference and convergence time for ResTT with different virtual bond dimensions on MNIST dataset are given in Table \ref{Tab:Time}. Both IT and BT are less than 1 second. The increase in $r$ leads to an increase in IT and BT. Meanwhile, CT decreases as $r$ increases, unless $r$ is too large.

Next, we conduct multiple sets of experiments by choosing $r$ in $\{20, 50, 100\}$ and varying the fraction of data for training by $\{1\%, 5\%, 10\%, 20\%,50\%, 100\% \}$. Each model has been run 10 times to average the effect of random initialization. The mean and standard deviation of the test accuracy are reported in Table \ref{Tab:Mresult}. When the fraction is $1 \% $, that is, only 600 randomly selected samples are used for training, ResTT can achieve an error rate of $10.18 \% $ on the test set which contains 10000 test samples.  Moreover, it is worth mentioning that ResTT converges much faster than MeanTT. As can be seen from \fref{Fig:Train}, the training of MeanTT with $r = 20$ is highly unstable on a fraction of data, while ResTT can converge to an approximately optimal point within just 20 epoches. In \fref{Fig:Train} (b), ResTT with $r = 100$ takes only about $5$ epoches to complete the training. According to the numerical results, ResTT consistently outperforms MeanTT in terms of the training stability and test accuracy, even if we have shown that MeanTT is already more stable in training than other tensor network models such as regularized MPS and TT-BN. Therefore, it is clear that the residual connections, with proper initialization guided by the mean-field analysis, can indeed improve the effectiveness and training stability of the model.

\subsection{Image classification with limited data}
To further study the performance of ResTT and benchmark deep learning models in scenarios with limited data, we conduct experiments using a small amount of training data from MNIST and Fashion-MNIST. 

\begin{table}[]
	\renewcommand\arraystretch{1.3}
	\centering
	\vspace{-5pt}
	\caption{Comparisons of the classification accuracy ($\%$) with different fractions of training data on the test sets of MNIST and Fashion-MNIST. The virtual bond dimension for ResTT is 100.}
	\label{Tab:result_limit}
	\begin{tabular}{c|c|c|c}
		\hline
	Fraction	    &Model& MNIST   & Fashion-MNIST \\ \hline
		\multirow{5}{*}{1\%}
		&LR     & 9.8$\pm$0 & 10$\pm$0 \\
		&MLP       &67.28$\pm$8.896 & 10$\pm$0\\
		&LeNet-5     & 84.59$\pm$6.781   & 75.90$\pm$0.8228 \\
		&ResNet18    & 75.44$\pm$1.011   & 75.66$\pm$0.3875 \\
		&ResTT    & \textbf{89.82$\pm$0.2755}  & \textbf{78.86$\pm$0.3120} \\
		\hline
		\multirow{5}{*}{5\%}&LR     & 9.8$\pm$0 & 10$\pm$0 \\
		&MLP       & 75.31 $\pm$10.03 & 10$\pm$0 \\
		&LeNet-5     & 93.61$\pm$1.527  &  80.52$\pm$0.8879\\
		&ResNet18    & 90.43$\pm$0.6631   &81.71$\pm$0.2287 \\
		&ResTT    & \textbf{95.46$\pm$0.0768}    &  \textbf{84.18$\pm$0.1554} \\
		\hline
		\multirow{5}{*}{10\%}&LR     & 9.8$\pm$0 &10$\pm$0 \\
		&MLP       & 76.33$\pm$5.195 & 10$\pm$0 \\
		&LeNet-5     & 94.71$\pm$1.431   & 82.34$\pm$0.7049 \\
		&ResNet18    & 94.84$\pm$0.3111  & 83.95$\pm$0.1931 \\
		&ResTT    &  \textbf{96.74$\pm$0.0706}    & \textbf{85.69$\pm$0.1116}  \\
		\hline
	\end{tabular}
\end{table}

\subsubsection{Datasets}
The fraction of training data for MNIST and Fashion-MNIST are selected from $\{1\%, 5\%, 10\%\}$. The training data are randomly sampled from the original training set. All experiment are repeated 10 times at each fraction to alleviate the effect of randomness.
\subsubsection{Models} Models for comparison are listed as follows:
\begin{itemize}
	\item [a)] \textit{LR:} Linear regression is a statistical model that captures the linear correlation of the input features.
	\item [b)] \textit{MLP:} Multi-layer perceptron is the feed-forward neural network that consists of multiple fully-connected layers (with threshold activation). With a layer-by-layer structure, MLP is able to distinguish data that is not linearly separable. The MLP here is adopted from \cite{xiao2017fashion}, which is composed of two layers with 100 hidden neurons and the ReLU activation.
	\item [c)] \textit{LeNet-5: } LeNet-5 is the light-weight convolutional neural network (CNN) that consists of two sets of convolutional and average pooling layers, followed by two fully-connected layers and finally a Softmax layer.
	\item [d)] \textit{ResNet18:} ResNet \cite{he2016deep} is composed of multiple residual blocks and is one of the most influential CNN structure in deep learning literature. The ResNet18 refers to the 18-layer deep ResNet.
\end{itemize}

\subsubsection{Results}
Results with various fractions of training data are shown in Table \ref{Tab:result_limit}. For LR and MLP, since the models are overly simple, limited training data may result in underfitting and non-convergence. It is clear that the performance of ResTT is significantly better than LeNet-5 and ResNet18 in terms of prediction accuracy and training robustness for different fractions. Particularly, for the fraction of 1\%, ResTT also beats the multi-task learning methods \cite{yang2017deep} on MNIST, including the Deep Multi-Task Representation Learning model based on Tensor Train (DMTRL-TT) ($\sim87\%$), DMTRL-Tucker ($\sim83\%$) and DMTRL-LAF ($\sim85\%$).

\subsection{Practical examples with limited data}
In this subsection, we demonstrate the application of ResTT on real tasks with limited and sequential data which are known to have complex feature interactions. Particularly, the experiments are conducted on Boston Housing and ALE datasets where ResTT are compared with other popular methods.
\subsubsection{Datasets}
\label{5-4-1}
\begin{itemize}
	\item [a)] \textit{Boston Housing:} The Boston Housing dataset is collected by the U.S. Census Service concerning housing in the area of Boston Mass. The task is to use 13 variables, such as crime rate and tax rate, to predict Boston housing prices. The dataset is small with only 506 samples. We use 354 samples (70\%) for training and the rest (152 samples) for testing. 
	
	\item [b)] \textit{ALE:} The dataset is to predict the average localization error (ALE) with applications to wireless sensor networks, which consists of four features, namely anchor ratio, transmission range of a sensor, node density and iteration count. The dataset is small with 107 samples. We use 74 samples (70\%) for training and the rest (33 samples) for testing. 
\end{itemize}

\subsubsection{Models} Models for comparison are listed as follows:
\begin{itemize}
	\item [a)] \textit{LR:} Linear regression (LR) is a statistical model that captures the linear correlation of the input features.
	\item [b)] \textit{PR:} Polynomial regression (PR) is a statistical model that captures the polynomial correlations of the input features. We use the quadratic model, which exploits quadratic interaction terms of the input features for modelling.
	\item [c)] \textit{RR:} Ridge regression (RR) is a special LR model constrained by L2 regularization.
	\item [d)] \textit{PRR:} Polynomial ridge regression (PRR) is a polynomial RR model. Here, we use the quadratic model.
	\item [e)] \textit{LaR:} Lasso regression (LaR) is a special LR model constrained by L1 regularization.
	\item [f)] \textit{PLaR:} Polynomial Lasso regression (PLaR) is a polynomial LaR model. Here, we use the quadratic model.
	\item [g)] \textit{SVR:} Support vector regression (SVR) is a kernel-based regression method, which aims to learn a regression hyperplane in high-dimensional feature space \cite{smola2004tutorial}. Here, we adopt the radial basis function (RBF) kernel.
	\item [h)] \textit{MLP:} The MLP here is composed of two layers with 100 hidden neurons and the ReLU activation.

\end{itemize}

\begin{table}[]
	\renewcommand\arraystretch{1.3}
	\centering
	\caption{Comparisons of R2 score and RMSE on the test sets of Boston Housing and ALE. The virtual bond dimension for ResTT is 50.}
	\label{Tab:Rresult}
	\begin{tabular}{c|c|c|c|c}
		\hline
		Model    &  \multicolumn{2}{c|}{Boston Housing}& \multicolumn{2}{c}{ALE} \\ \hline
		             &R2 score&RMSE&R2 score&RMSE \\ \hline
		LR          & 0.654 & 4.61&0.715& 0.181    \\\hline
		PR          & 0.779 & 3.68	&0.591 & 0.218     \\\hline
		RR          & 0.664 & 4.55&0.719  & 0.183  \\\hline
		PRR          & 0.831 & 3.23&0.593 & 0.216    \\\hline
		LaR          & 0.673& 4.48&0.721& 0.182 \\\hline
		PLaR          & 0.848 & 3.05&0.619& 0.209   \\\hline
		SVR          & 0.846 & 3.07&0.644 & 0.202  \\\hline
		MLP          & 0.829 & 3.47&0.704  & 0.201  \\\hline
		ResTT          & \textbf{0.864} & \textbf{2.96}&\textbf{0.736}  & \textbf{0.179}     \\\hline
	\end{tabular}
\end{table}

\subsubsection{Results}
We adopt the RMSE and R2 score \cite{freedman2009statistical} as the evaluation metrics. Experimental results are shown in Table \ref{Tab:Rresult}. We see that ResTT achieves the best performance among all the competitors. For the Boston Housing dataset, the performance of polynomial models is significantly better than that of linear models, whereas the linear models outperform the polynomial models on the ALE dataset. Note that ResTT achieves stable performance on both datasets, which demonstrates its effectiveness for modelling multiple multilinear correlations.

\section{Conclusion}
\label{6}
In this paper, we propose a general framework that combines residual connections with TT to model feature interactions, from $1$-order to $N$-order, within one model. Mean-field analysis has been extended to TT and ResTT to derive the condition of stability for the initialization of deep tensor networks. In particular, the stability condition of ResTT is significantly relaxed due to the residual connections. ResTT has demonstrated significant improvement in training stability and convergence speed over the plain TT both theoretically and experimentally.

Numerical experiments on the synthetic, image classification datasets and two practical tasks have shown the robust performance of ResTT. Particularly, ResTT outperforms the benchmark deep learning models in the case of limited training data.

The future direction would be to evaluate these tools on more complex problems in which high-order multi-modal correlations naturally exist, such as interpreting electroencephalography signals \cite{zhang2019making}. For example, it would be interesting to explore the application of ResTT in multi-task learning and continual learning models, for which the high-order feature interaction is often adopted. As ResTT can be seen as an enhanced version of Volterra series, we also anticipate its applications in the learning with sequential data \cite{zhang2019making}. Moreover, considering the excellent performance of ResTT on limited training data, it is promising to apply ResTT to tasks for which training data acquisition is expensive \cite{chen2019semisupervised}.

\begin{appendices}
	\section{Detailed Expression of the Output of a General ResTT}
	\label{Ap.1}
	The elements of the final output of the ResTT with $N$ input features are calculated in the following, ranked from high-order to low-order. Firstly, the $N$-order term is calculated to be
	\begin{align}
	\sum_{v_1,\cdots,v_{N-1}} \sum_{i_1,\cdots,i_N} w^{(1,1)}_{i_{1}v_{1}} w^{(2,1)}_{v_{1}i_{2}v_{2}} \cdots  w^{(N,1)}_{v_{N-1}i_{N}}  x^{(1)}_{i_1} \cdots x^{(N)}_{i_N}.\nonumber\\
	\end{align}
This term is the output of a plain TT network. Secondly, the ($N-1$)-order terms are calculated to be
	\begin{eqnarray}
	&\sum_{v_2,\cdots,v_{N-1}} \sum_{i_2,\cdots,i_N} w^{(2,2)}_{i_{2}v_{2}} \cdots  w^{(N,1)}_{v_{N-1}i_{N}}&\nonumber\\
    &x^{(2)}_{i_2} \cdots x^{(N)}_{i_N},& \nonumber\\
	&\cdots\cdots&   \nonumber\\
	&\sum_{v_{1},\cdots,v_{k-1},v_{k+1},\cdots,v_{N-1}} \sum_{i_1,\cdots,i_{k-1},i_{k+1},\cdots,i_N}&\nonumber\\
    &w^{(1)}_{i_{1}v_{1}}\cdots w^{(k-1,2)}_{i_{k-1}v_{k-1}}  w^{(k+1,1)}_{v_{k-1}i_{k+1}v_{k+1}} \cdots w^{(N)}_{v_{N-1}i_{N}}&\nonumber\\
    &x^{(1)}_{i_1} \cdots x^{(k-1)}_{i_{k-1}} x^{(k+1)}_{i_{k+1}}\cdots x^{(N)}_{i_N},& \nonumber\\
	&\cdots\cdots& \nonumber\\
	&\sum_{v_{1},\cdots,v_{N-2}} \sum_{i_1,\cdots,i_{N-1}} w^{(1,1)}_{i_{1}v_{1}}\cdots w^{(N,3)}_{v_{N-1}v_{N}}&   \nonumber\\
	&x^{(1)}_{i_1} \cdots x^{(N-1)}_{i_{N-1}}.&
	\end{eqnarray}
Similarly, the ($N-2$)-order terms can be calculated as
	\begin{eqnarray}
	&\sum_{v_3,\cdots,v_{N-1}} \sum_{i_3,\cdots,i_N} w^{(3,2)}_{i_{3}v_{3}} w^{(4,1)}_{v_{3}i_{4}v_{4}} \cdots  w^{(N)}_{v_{N-1}i_{N}v_{N}}&\nonumber\\
&x^{(3)}_{i_3} \cdots   x^{(N)}_{i_N},&\nonumber\\
    &\cdots\cdots&  \nonumber\\	
	&\sum_{v_{1},\cdots,v_{k-1},v_{k+2},\cdots,v_{N-1}} \sum_{i_1,\cdots,i_{k-1},i_{k+2},\cdots,i_N}&\nonumber\\
    &w^{(1)}_{i_{1}v_{1}} \cdots w^{(k-1,2)}_{i_{k-1}v_{k-1}}w^{(k+2,1)}_{v_{k-1}i_{k+2}v_{k+2}} \cdots w^{(N,1)}_{v_{N-1}i_{N}v_{N}}&\nonumber\\
    &x^{(1)}_{i_1} \cdots x^{(k-1)}_{i_{k-1}} x^{(k+2)}_{i_{k+2}} \cdots x^{(N)}_{i_N},& \nonumber\\
	&\cdots\cdots&   \nonumber\\
	&\sum_{v_{1},\cdots,v_{N-2}} \sum_{i_1,\cdots,i_{N-2}} w^{(1,1)}_{i_{1}v_{1}}\cdots w^{(N,3)}_{v_{N-2}v_{N}}&  \nonumber\\
	&x^{(1)}_{i_1} \cdots x^{(N-2)}_{i_{N-2}}.&
	\end{eqnarray}
The rest of the interaction terms can be obtained in the same manner. Finally, we have the following 1-order terms (linear correlations)
	\begin{align}
	&\sum\limits_{i_1}w^{(1,1)}_{i_{1}v_N}x^{(1)}_{i_{1}}
	,\sum\limits_{i_2}w^{(2,2)}_{i_2 v_N}x^{(2)}_{i_{2}},\cdots,\sum\limits_{i_N}w^{(N,2)}_{i_N v_N}x^{(N)}_{i_{N}}.
	\end{align}
The final output of ResTT is the summation of all terms, from 1-order (linear) to $N$-order ($N$-linear).

	\section{Mean-field Analysis of TT and ResTT}
	\label{Ap.2}
Recall that all the weights are initialized from the same Gaussian distribution $\mathcal N(0,\sigma^{2}_{w}/r)$. We have
	\begin{equation}
	{\mathbb E}(y^{(k)}_{v_{k}}) = 0.
	\end{equation}
Then the variance of the $k$-th output vector is
	\begin{align}
	q^{(k)} = {\mathbb E}[(\sum_{v_{k-1},i_{k}}y^{(k-1)}_{v_{k-1}}w^{(k)}_{v_{k-1}i_{k}v_{k}}x^{(k)}_{i_{k}})^2].
	\label{Eq:appendix-qk}
	\end{align}
Since $y^{(k-1)}_{v_{k-1}}, w^{(k)}_{v_{k-1}i_{k}v_{k}}, x^{(k)}_{i_{k}}$ are independent, we have
	\begin{align}
	{\mathbb E}[y^{(k-1)}_{v_{k-1}} w^{(k)}_{v_{k-1}i_{k}v_{k}} x^{(k)}_{i_{k}}]  = 0.
	\end{align}
	Hence, \eref{Eq:appendix-qk} can be simplified as
	\begin{align}
	q^{(k)} & ={\mathbb E}[\sum_{v_{k-1},i_{k}}(y^{(k-1)}_{v_{k-1}})^2(w^{(k)}_{v_{k-1}i_{k}v_{k}})^2(x^{(k)}_{i_{k}})^2] \nonumber\\
	&={\mathbb E}[\sum_{i_{k}}(x^{(k)}_{i_{k}})^2)]\sigma^{2}_{w}q^{(k-1)}.
	\end{align}
	Similarly, the derivations of quadratic mean $\chi^{(k)}$ of the elements in $\mathcal{J}^{(k)}$ are given by
	\begin{align}
	\chi^{(k)}&=\frac{1}{r}{\mathbb E}(\sum_{v_{k-1}}{(j_{v_{k-1}}^{(k)})^2})=\frac{1}{r}{\mathbb E}(\sum_{v_{k-1}}(j^{(k+1)}\cdot a^{(k)})^2_{v_{k-1}}) \nonumber\\
	&=\frac{1}{r}{\mathbb E}[\sum_{v_{k},v_{k-1}}(j^{(k+1)}_{v_k})^2(a^{(k)}_{v_kv_{k-1}})^2]\nonumber\\
	&=\frac{1}{r}{\mathbb E}[\sum_{v_{k},v_{k-1},i_{k}}(j^{(k+1)}_{v_k})^2(w^{(k)}_{v_{k}i_{k}v_{k-1}})^2(x^{(k)}_{i_{k}})^2]\nonumber\\
	&= {\mathbb E}(\sum_{i_{k}}(x^{(k)}_{i_{k}})^2)\sigma^{2}_{w}\chi^{(k+1)}
	\end{align}
The evolution of signals in ResTT is slightly different due to the residual terms. Given that
	\begin{equation}
	\mathcal{Y}^{(k)} = \mathcal{Y}^{(k,1)} +  \mathcal{Y}^{(k,2)} + \mathcal{Y}^{(k-1)} ,
	\end{equation}
we have
	\begin{align}
	q^{(k)} &= \frac{1}{r} {\mathbb E}(\sum_{v_{k},v_{k-1},i_{k}}(y^{(k-1)}_{v_{k-1}})^2(w^{(k,1)}_{v_{k-1}i_{k}v_{k}})^2(x^{(k)}_{i_{k}})^2) \nonumber\\
	&+ \frac{1}{r} {\mathbb E} (\sum_{i_{k},v_{k}}(x^{(k)}_{i_{k}})^2(w^{(k,2)}_{i_{k}v_{k}})^2) + \frac{1}{r} {\mathbb E} (\sum_{v_{k-1}}(y^{(k-1)}_{v_{k-1}})^2)\nonumber\\
	&=(\sigma^{2}_{w}{\mathbb E}(\sum_{i_{k}}(x^{(k)}_{i_{k}})^2)+1)q^{(k-1)} + \sigma^{2}_{w}{\mathbb E}(\sum_{i_{k}}(x^{(k)}_{i_{k}})^2) \nonumber\\
	&=(s^{(k)}+1)q^{(k-1)}+ s^{(k)},
	\end{align}
	and
	\begin{align}
	\chi^{(k)}=&\frac{1}{r}{\mathbb E}(\sum_{v_{k},v_{k-1}}(j^{(k)}_{v_k})^2(a^{(k)}_{v_kv_{k-1}} + 1)^2)\nonumber\\
	=&\frac{1}{r}{\mathbb E}(\sum_{v_{k},v_{k-1},i_{k}}(j^{(k+1,1)}_{v_k})^2 [ (w^{(k,1)}_{v_{k}i_{k}v_{k-1}})^2(x^{(k)}_{i_{k}})^2 + 1])\nonumber\\
	=& (s^{(k)}+1)\chi^{(k+1)}.
	\end{align}
\end{appendices}

\bibliographystyle{IEEEtran} 			 
\bibliography{ref}                        

\end{document}